\documentclass[letterpaper]{article} % DO NOT CHANGE THIS
\usepackage{aaai25}  % DO NOT CHANGE THIS
\usepackage{times}  % DO NOT CHANGE THIS
\usepackage{helvet}  % DO NOT CHANGE THIS
\usepackage{courier}  % DO NOT CHANGE THIS
\usepackage[hyphens]{url}  % DO NOT CHANGE THIS
\usepackage{graphicx} % DO NOT CHANGE THIS
\usepackage{natbib}  % DO NOT CHANGE THIS AND DO NOT ADD ANY OPTIONS TO IT
\usepackage{caption} % DO NOT CHANGE THIS AND DO NOT ADD ANY OPTIONS TO IT
\usepackage{algorithm}
\usepackage{algorithmic}
 \usepackage{amsmath}
\usepackage{newfloat}
\usepackage{listings}
\usepackage{enumitem}
\usepackage{booktabs} 
\usepackage{multirow} 
\usepackage{booktabs} 
\usepackage{amssymb}
\DeclareCaptionStyle{ruled}{labelfont=normalfont,labelsep=colon,strut=off} % DO NOT CHANGE THIS
\floatstyle{ruled}
\newfloat{listing}{tb}{lst}{}
\floatname{listing}{Listing}
\title{IteRPrimE: Zero-shot Referring Image Segmentation with Iterative Grad-CAM Refinement and Primary Word Emphasis}
\author{
    Yuji Wang\textsuperscript{*}, Jingchen Ni\textsuperscript{*}, Yong Liu, Chun Yuan, Yansong Tang\textsuperscript{†}
}
\affiliations{
    Shenzhen International Graduate School, Tsinghua University\\
    \textsuperscript{*}\{yuji-wan24, njc24\}@mails.tsinghua.edu.cn, \textsuperscript{†}tang.yansong@sz.tsinghua.edu.cn
}

\usepackage{bibentry}
\begin{document}

\maketitle

\begin{abstract}
Zero-shot Referring Image Segmentation (RIS) identifies the instance mask that best aligns with a specified referring expression without training and fine-tuning, significantly reducing the labor-intensive annotation process. Despite achieving commendable results, previous CLIP-based models have a critical drawback: the models exhibit a notable reduction in their capacity to discern relative spatial relationships of objects. This is because they generate all possible masks on an image and evaluate each masked region for similarity to the given expression, often resulting in decreased sensitivity to direct positional clues in text inputs. Moreover, most methods have weak abilities to manage relationships between primary words and their contexts, causing confusion and reduced accuracy in identifying the correct target region. To address these challenges, we propose \textbf{IteRPrimE} (\textbf{Ite}rative Grad-CAM \textbf{R}efinement and \textbf{Prim}ary word \textbf{E}mphasis), which leverages a saliency heatmap through Grad-CAM from a Vision-Language Pre-trained (VLP) model for image-text matching. An iterative Grad-CAM refinement strategy is introduced to progressively enhance the model's focus on the target region and overcome positional insensitivity, creating a self-correcting effect. Additionally, we design the Primary Word Emphasis module to help the model handle complex semantic relations, enhancing its ability to attend to the intended object. Extensive experiments conducted on the RefCOCO/+/g, and PhraseCut benchmarks demonstrate that IteRPrimE outperforms previous SOTA zero-shot methods, particularly excelling in out-of-domain scenarios. 

% Code will be available: https://github.com/VoyageWang/IteRPrimE.
% \link{Datasets}{https://aaai.org/example/datasets}
% \link{Extended version}{https://aaai.org/example/extended-
\end{abstract}
\begin{links}
\link{Code}{https://github.com/VoyageWang/IteRPrimE}
\end{links}

\section{Introduction}

Referring Image Segmentation (RIS) requires the model to generate a pixel-level referred object mask based on a textual description, extending the applicability to various tasks such as robot interaction and image editing \cite{TPAMILAVT,lavt, unilseg, lisa, soc}. Different from standard semantic segmentation \cite{ERSS,wang2024convolution, ss-1,iteross, bai2024self}, RIS necessitates the differentiation of instances within the same category and their relationships with other objects or the scene, which requires high demands on the semantic understanding and spatial perception of the model. However, annotating exact pairs of images, descriptions, and ground-truth masks is both expensive and time-intensive, as the annotation of a query needs a grasp of diverse positional and attributive details within the image \cite{gres, mevis,clevr}. Recent weakly supervised RIS techniques \cite{tseg, chunck,groupvit} have been introduced to mitigate these annotation challenges, yet they still depend on paired data for training purposes and have relatively poor performance. In contrast, a zero-shot approach holds greater value. Leveraging vision-language pre-trained (VLP) models such as CLIP \cite{CLIP}, this method efficiently generalizes across diverse concepts and unseen categories without further training and fine-tuning.
\begin{figure*}
	\centering
	\includegraphics[width=0.9\textwidth]{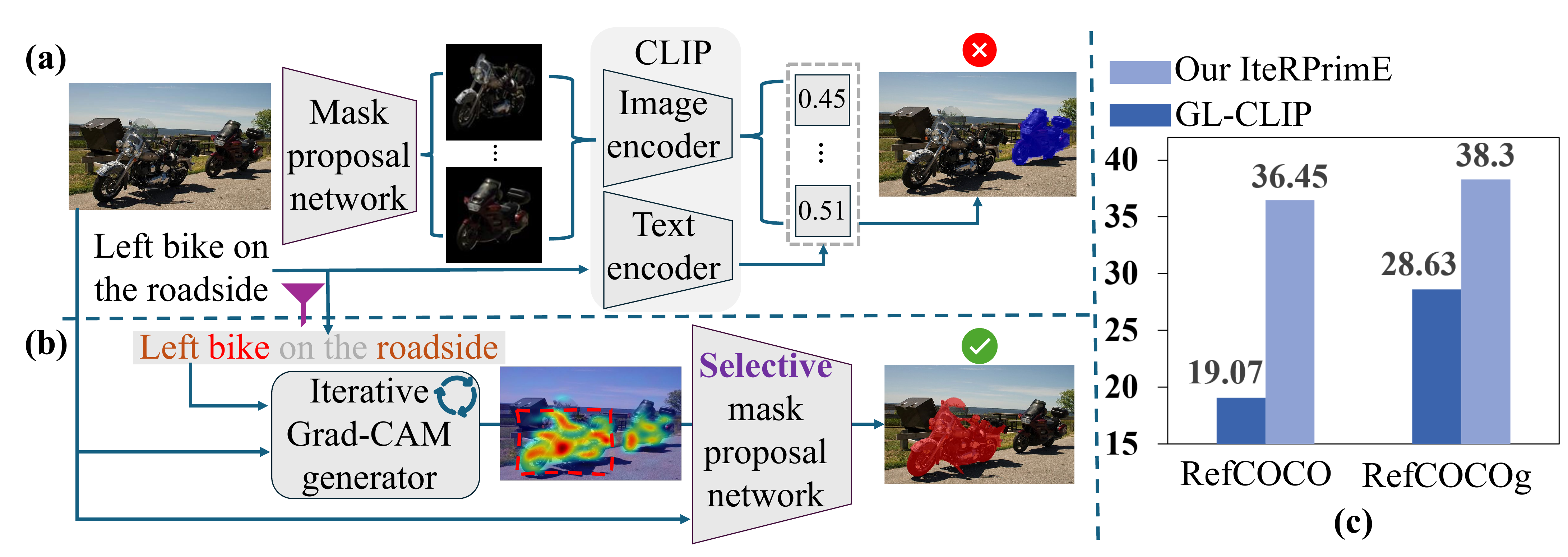}
	\caption{(a) The general pipeline of CLIP-based methods. They lack the perception of spatial relative position due to the masked images. (b) The pipeline of our IteRPrimE with Iterative Grad-CAM Refinement Strategy and Primary Word Emphasis of ``bike". (c) This is a comparative experiment of positional phrase accuracy between IteRPrimE and GL-CLIP on RefCOCO and RefCOCOg.
 }
	\label{fig1}
\end{figure*}

Existing methodologies to harness the characteristics of being unnecessary to fit training data presented by zero-shot learning often employ a two-stage pipeline, shown in Figure \ref{fig1} (a). As a discriminator between the images masked by the candidate masks and the expression, CLIP is used to select the instance mask whose similarity score is the highest~\cite{car, gcclip, tas, refdiff}. However, we observed that these methods always malfunctioned when encountering text inputs with positional information such as ``left" and ``right". Due to only a single instance contained in a masked image, the absence of relative spatial perception can be the inherent limitation of these CLIP-based paradigms. Previous pieces of literature alleviate this issue by injecting the human priors or bias that explicitly prompts the CLIP with the given direction clues \cite{refdiff, tas}. To be more specific, they manually design spatial decaying weights from 1 to 0 in the directions consistent with text phrases to make the model aware of positional information, but it can not generalize the scenarios out of predefined directions such as ``next to".
Additionally, the domain shift for CLIP from the natural image to the masked image can also impact the segmentation performance \cite{ovs_liu, zegformer, survey}. 
% % 这里可以准备两个版本，第一就是 masked image和 第二个就是这个

\begin{figure}
	\centering
	\includegraphics[width=0.47\textwidth]{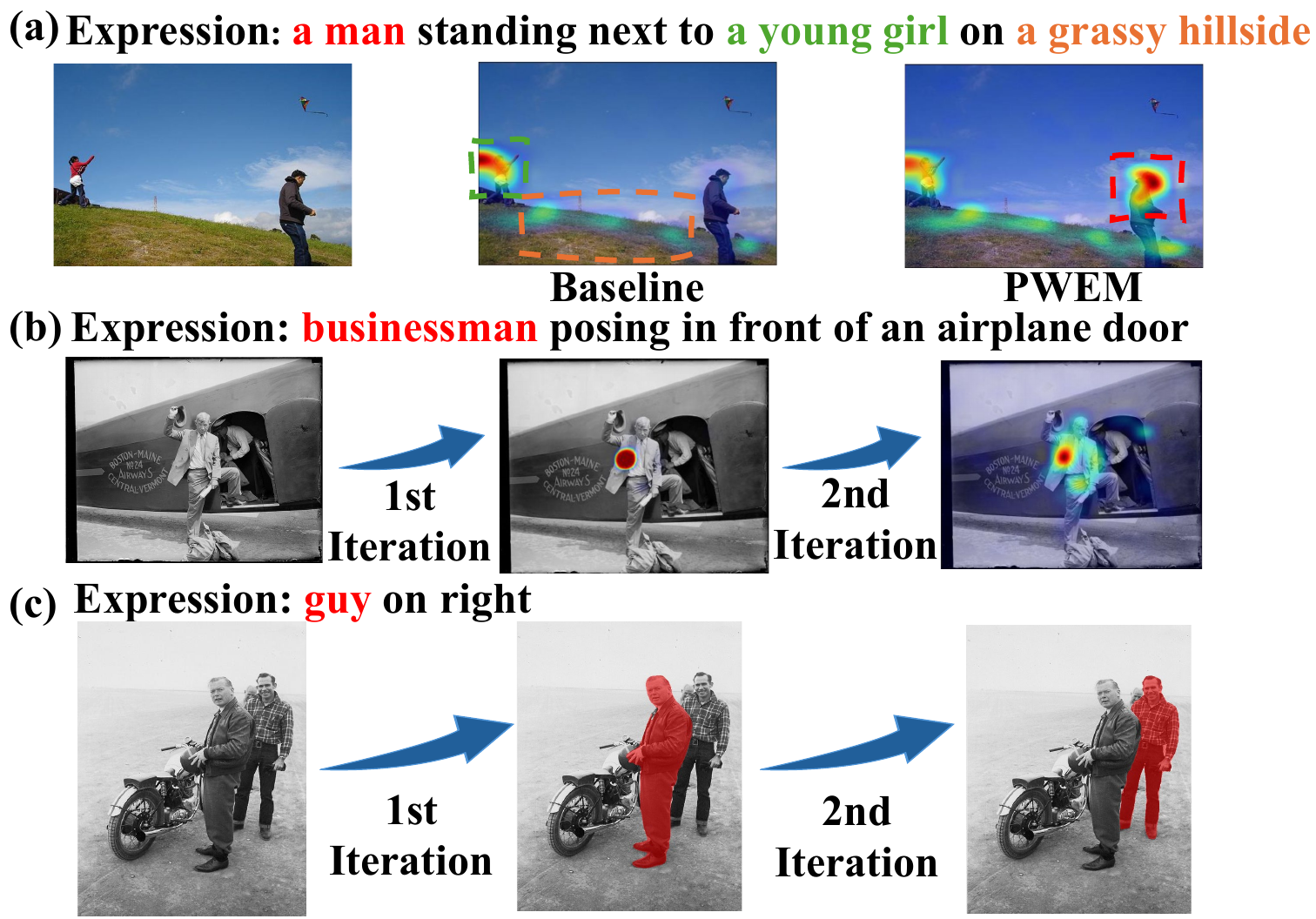}
	\caption{(a) The weak ability of the baseline model to differentiate the semantic relationships between the primary word ``man" and the other noun phrases colored green and orange. PWEM can make the model aware of the targeted instance referred to by the main word. (b) The IGRS facilitates the expansion of highlighted areas, surpassing the confined small regions. (c) IGRS offers the model chances of self-correction. }
	\label{fig2}
\end{figure}

% 写作技巧，需不需要在这里就把他说了 We propose the iterative strategy to 
Some researchers \cite{chunck} have leveraged Grad-CAM \cite{gradcam} and created two specialized loss functions to attenuate the detrimental effects of positional phrases in weakly supervised settings. Although the losses are unsuitable for zero-shot scenarios, Grad-CAM can partially mitigate the deleterious effects associated with masked images. This is because the method maintains the integrity of the model’s spatial perception capabilities by delineating the regions with the greatest attention in the original image for localization, shown in Figure \ref{fig1} (b). Nevertheless, we still find two major problems by analyzing the occurrences and characteristics of Grad-CAM. First, Grad-CAM struggles to discriminate the semantic relations between different noun phrases, due to the lack of a stronger consideration of the primary word than other context words, shown in baseline predictions of Figure \ref{fig2} (a). Specifically, the model’s weak ability to effectively prioritize the main word in complex expressions undermines its overall performance. Second, Grad-CAM is limited to identifying only small areas of the referred object, which consequently results in selecting undesired instance masks.

To overcome these challenges, we propose a novel framework namely, \textbf{IteRPrimE} (\textbf{Ite}rative Grad-CAM \textbf{R}efinement and \textbf{Prim}ary word \textbf{E}mphasis) utilizing Grad-CAM for zero-shot RIS. First, we implement an iterative refinement strategy to enhance the representational accuracy and enlarge the indicated area of Grad-CAM, progressively improving the model's concentration on the target object with each cycle, shown in Figure \ref{fig2} (b). Simultaneously, this strategy is particularly beneficial when the referring expression includes positional words, as it offers the model chances of self-correction at each iteration, shown in Figure \ref{fig2} (c). Second, the Primary Word Emphasis Module (PWEM) plays a crucial role in enhancing the weak abilities to handle the complex semantic relationships between primary words and other contexts. This module is achieved by emphasizing the Grad-CAMs of the main word within the referring expression, from local and global aspects. Finally, a post-processing module is designed to select a high-quality, contiguous instance mask from a mask proposal network, which encapsulates the target object as indicated by Grad-CAM. By addressing the limitations, the IteRPrimE approach achieves superior performance over prior zero-shot state-of-the-art techniques, notably excelling in out-of-domain scenarios and exhibiting robust cross-domain transfer proficiency. Our main contributions include 
\begin{enumerate}

	\item  To our best knowledge, we are the first to use Grad-CAM to instruct Segmentors for zero-shot RIS tasks. 
 \item We propose the Iterative Grad-CAM Refinement Strategy (IGRS) and Primary Word Emphasis Module (PWEM) to enhance the accuracy and representation of Grad-CAM for better localization, shown in Figure \ref{fig2}. 
	\item Compared to the previous CLIP-based method, our method significantly outperforms it with inputs containing positional information,  shown in Figure \ref{fig1} (c). Additionally, the approach achieves a better performance on the four popular benchmarks, especially for the out-domain datasets. 
\end{enumerate}

\section{Related Works}

\textbf{Zero-shot referring image segmentation.} For the fully-supervised setting, training a well-specialized model for RIS needs massive paired text-visual annotations, which are sometimes not affordable and accessible \cite{lqmformer, gres, lavt, cris, restr, ris-1,ris-2}. Besides, these models have relatively weak ability in out-of-domain scenarios due to the limited data and a domain gap. Therefore, the zero-shot RIS methods are proposed as the alternative. Global- and local-CLIP (GL-CLIP) \cite{gcclip} is the first proposed to segment the instance given the text input with zero-shot transfer. By interfacing with the mask proposal network FreeSOLO \cite{freesolo}, the approach leverages both global and local textual-image similarity to enhance the discriminative capabilities of the CLIP model. Based on GL-CLIP,  some researchers \cite{bsap} combine the original CLIP similarity score with their proposed Balanced Score with Auxiliary Prompts (BSAP), namely BSAP-H, to reduce the CLIP’s text-to-image retrieval hallucination. Ref-Diff \cite{refdiff} demonstrates that the text-to-image generative model like Stable Diffusion \cite{sd} can generate the intended mask from the cross-attention map, which has considerable performance. TAS \cite{tas} mainly depends on another large captioner network BLIP2 \cite{blip2} to mine the negative text based on the previous mask proposal network plus discriminator paradigm, which achieves favorable performances. Additionally, SAM \cite{sam} is utilized for better segmentation accuracy. However, these CLIP-based methods struggle to segment the referred subject with positional-described text queries, due to the absence of spatial relationships in the masked image. 

\textbf{Grad-CAM for localization.} Grad-CAM \cite{gradcam} is proposed to provide explainable clues indicating the regions the model pays attention to for the prediction head. In the context of the Image Text Matching (ITM) objective from any VLP \cite{vlp1, vlp-2, vlp-3}, Grad-CAM enables the establishment of a modality mapping from the textual to the visual domain, specifically calibrated for the task of visual localization. Many works utilize it to localize the objects with the given text \cite{groundvlp, chunck, groupvit, vlmae, ALBEF}. However, these approaches either generate a bounding box annotation or are employed within weakly supervised scenarios. Compared to approaches \cite{reco, extract, iteross} that perform zero-shot open vocabulary semantic segmentation with Grad-CAM, we are the first to propose the Grad-CAM for zero-shot RIS to study its behaviors under longer and complex textual inputs instead of a single category noun. To address problems of lacking consideration between main words and the other, the PWEM is proposed to aggregate the Grad-CAM from local-spatial and global-token levels. Secondly, a novel iterative refinement strategy is employed to obtain a better representation of Grad-CAM step by step.

\begin{figure*}
	\centering	\includegraphics[width=0.84\textwidth]{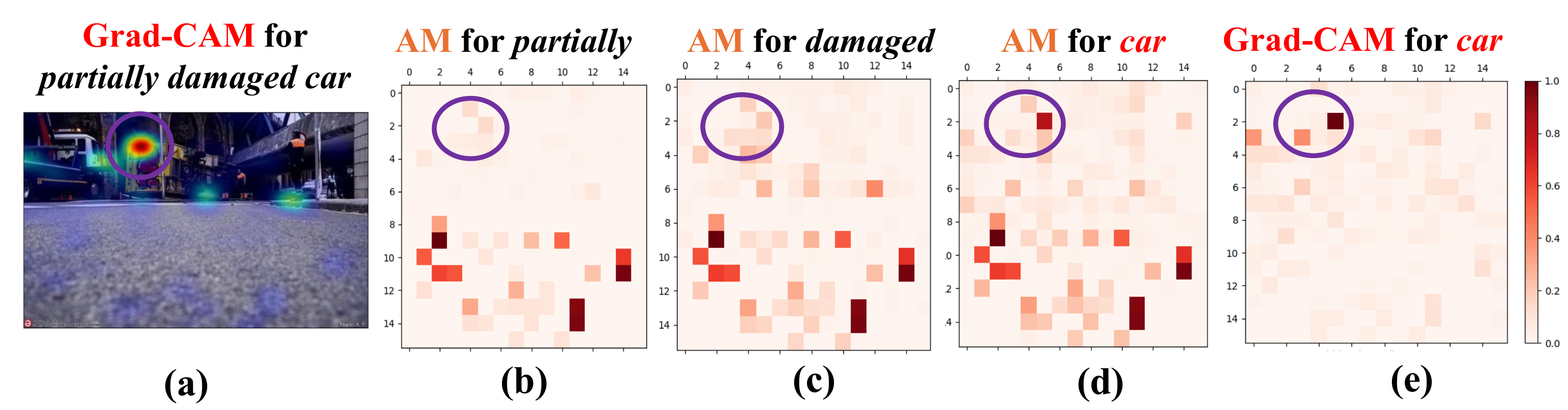}
	\caption{The Grad-CAMs and attention maps (AM) of ``partially damaged car''. Since the attention map (d) and Grad-CAM (e) of the primary word ``car'' both contain unique activation areas compared to the others, they can be harnessed from local-spatial and global-token perspectives to enhance the focus on the targeted regions, respectively. }
	\label{fig:main word}
\end{figure*}
\section{Preliminaries}
The generation of Grad-CAM is essential for harnessing it for RIS. Given an image-expression pair $(I, E)$, we can obtain their corresponding embeddings, $v$ and $e$, by the visual encoder $v = f_I(I)$ and text encoder $e = f_T(E)$, respectively. Then, for multimodal fusion, these two embeddings are fed to the cross-attention layers used to align the visual and textual information \cite{coca, transformer}. The resultant attention activation maps, $\mathbf{A}$, can indicate the activated and recognized regions of $v$ concerning each query textual token in $e$. However, these indication clues are usually scattered and not densely distributed in the relevant regions. Thus, the gradients, $\mathbf{G}$ can be used to sharpen and dilute the effect of non-relevant regions in $\mathbf{A}$, where contribute less to the output objective, $y$, like Image Text Matching (ITM). The result of this gradient-weighted dilution process is known as Grad-CAM, $\mathbf{H}$. 

In the cross-attention layer, the Grad-CAM can be formulated by Equation (\ref{eq1}) 
\begin{subequations}\label{eq1}  
  \begin{gather}  
    \mathbf{H} = \mathbf{A} \odot \mathbf{G}, \label{eq1a} \\ 
    \mathbf{G} = {clamp} \left( \frac{\partial y}{\partial \mathbf{A}}, 0, \infty \right),  \label{eq1b} 
  \end{gather}  
\end{subequations}
where $clamp$ removes negative gradients, which often represent noise or irrelevant features. Finally, the Grad-CAM used to indicate the image regions, $\mathbf{H}_{f}$, is the mean over all the number of text tokens $|e|$, as shown in Equation (\ref{eq2})
\begin{equation}\label{eq2}
    \mathbf{H_{\textit{\text{f}}}} = \mathbb{E}_k \left( \mathbf{{H}}^{k} \right), k \in |e|,  \mathbf{H}_{f} \in {{\mathbb{R}}^{B \times h \times w}}
\end{equation}
where $\mathbf{{H}}^k$ denotes the Grad-CAM for the $k$-th text token, $B$ is the batch size, and $h\times w$ is the size of visual latent space. This averaging process treats every word equally and ignores the importance of the primary word, thereby undermining the performance of RIS.

\section{Method}
\subsection{Overview}
Figure \ref{fig1} (b) demonstrates the entire workflow of our method for zero-shot RIS, IteRPrimE, which can be divided into two parts: an iterative Grad-CAM generator and a selective mask proposal network. First,  the Grad-CAM generator is a VLP model with cross-attention layers. The proposed IGRS and PWEM are integrated into the generator. Finally, within the mask proposal network, a post-processing module is designed to select the candidate instance masks, ensuring the accurate and detailed localization of the target object.

\subsection{Primary Word Emphasis Module}
The PWEM is an essential component of the IteRPrimE, designed to confront the challenge posed by the weak capability of Grad-CAM to manage the semantic relationships in input texts featuring multiple potential referred nouns. This module emphasizes the Grad-CAM of the primary word in the expression, thereby increasing the focus on the main word during the averaging operation. Specifically, we first use an NLP processing toolbox to parse the part-of-speech (POS) tags of each word, filtering out a set of text tokens that includes special $<CLS>$ token of BERT \cite{bert}, nouns, adjectives, verbs, proper nouns, and numerals. These words are recognized as effective tokens $W$ that can provide distinct semantics and their contextual information. They are composed of primary words $W_m$ and their contexts $W_{c}$, where $W =   W_m \cup W_{c}$ and $ W_{m} \cap W_{c} = \emptyset  $. Then, we extract the primary noun from these effective words (e.g. ``car" in ``partially damaged car" shown in Figure \ref{fig:main word}) by employing a designed algorithm. It first generates a syntax tree, identifies the leftmost noun phrase (NP), and then finds the rightmost noun (NN) within that NP, which can be detailed in Algorithm 1 in the appendix.

As shown in the right part of Figure \ref{fig3}, we emphasize the effect of the primary word Grad-CAM from two perspectives: local spatial-level and global token-level augmentation. Different from the other contextual effective words $W_{c}$, the attention map  $\mathbf{A}^{W_m}$ and Grad-CAM $\mathbf{H}_{m}$ of the primary token holds the unique activated areas that probably indicate the correct localization of Grad-CAM shown in Figure \ref{fig:main word}. Therefore, to highlight and isolate the specific contribution of the primary word from the local spatial level, we compute the L\textsubscript{2} normalized differences, $\mathbf{A}_{{dif}}$ between the main word activation map and the other context word activation maps, $\mathbf{A}^{W_{c}}$. The activation difference is further integrated with gradients from the main word $\mathbf{G}^{W_{m}}$, forming a spatial modulator to indicate the local spatial importance in the main word Grad-CAM, $\mathbf{H}_{m}$. Thus, we can obtain the local spatial-level enhanced Grad-CAM of the primary word, $\mathbf{H}_{{l}}$, as shown in Equation (\ref{eq3})
\begin{subequations}\label{eq3}
  \begin{align}
   & \mathbf{A}_{{dif}} = \frac{{{\mathbf{A}^{{W_m}}} - {\mathbf{A}^{{W_c}}}}}{{||{\mathbf{A}^{{W_m}}} - {\mathbf{A}^{{W_c}}}|{|_2}}},\\
     & \mathbf{H}_{{l}} = {\mathbf{A}}_{{dif}} \odot \mathbf{G}^{W_m} \odot \mathbf{H}_{{m}},
  \end{align}
\end{subequations}
where $\mathbf{H}_{m} = \mathbf{A}^{W_m} \odot \mathbf{G}^{W_m}$ following Equation (\ref{eq1}). Broadcasting occurs when the dimensions do not match. 

From the global aspect, we manually add the weight of the main word Grad-CAM $\mathbf{H}_{{m}}$ along the token axis during mean operations, which provides additional enhanced focus on the primary token. Therefore, we can obtain the global token-level Grad-CAM ${\mathbf{H}_{{g}}}$ by Equation (\ref{eq4})
\begin{subequations}\label{eq4}
  \begin{align}
   & {W^{'}} = {W} \cup \{ {W_m}\}  \times {N_c},\\
     & {\mathbf{H}_g} = {\mathbf{A}^{{W^{'}}}} \odot {\mathbf{G}^{{W^{'}}}}
  \end{align}
\end{subequations}
where $N_c$ is the number of context tokens and $\{W_m\} \times N_c$ means repeating the main word for $N_c$ times. Finally, the resulting augmented Grad-CAM, ${\mathbf{H}}_{{a}}$, is the mean of concatenated local and global Grad-CAMs , $\mathbf{H}_{c}$, along the token axis, where $\mathbf{H}_{c}  = [{\mathbf{H}_g},{\mathbf{H}_l}]$. This map significantly improves the model's Grad-CAM localization accuracy, shown in PWEM of Figure \ref{fig2} (a).

\begin{figure*}
	\centering	\includegraphics[width=0.9\textwidth]{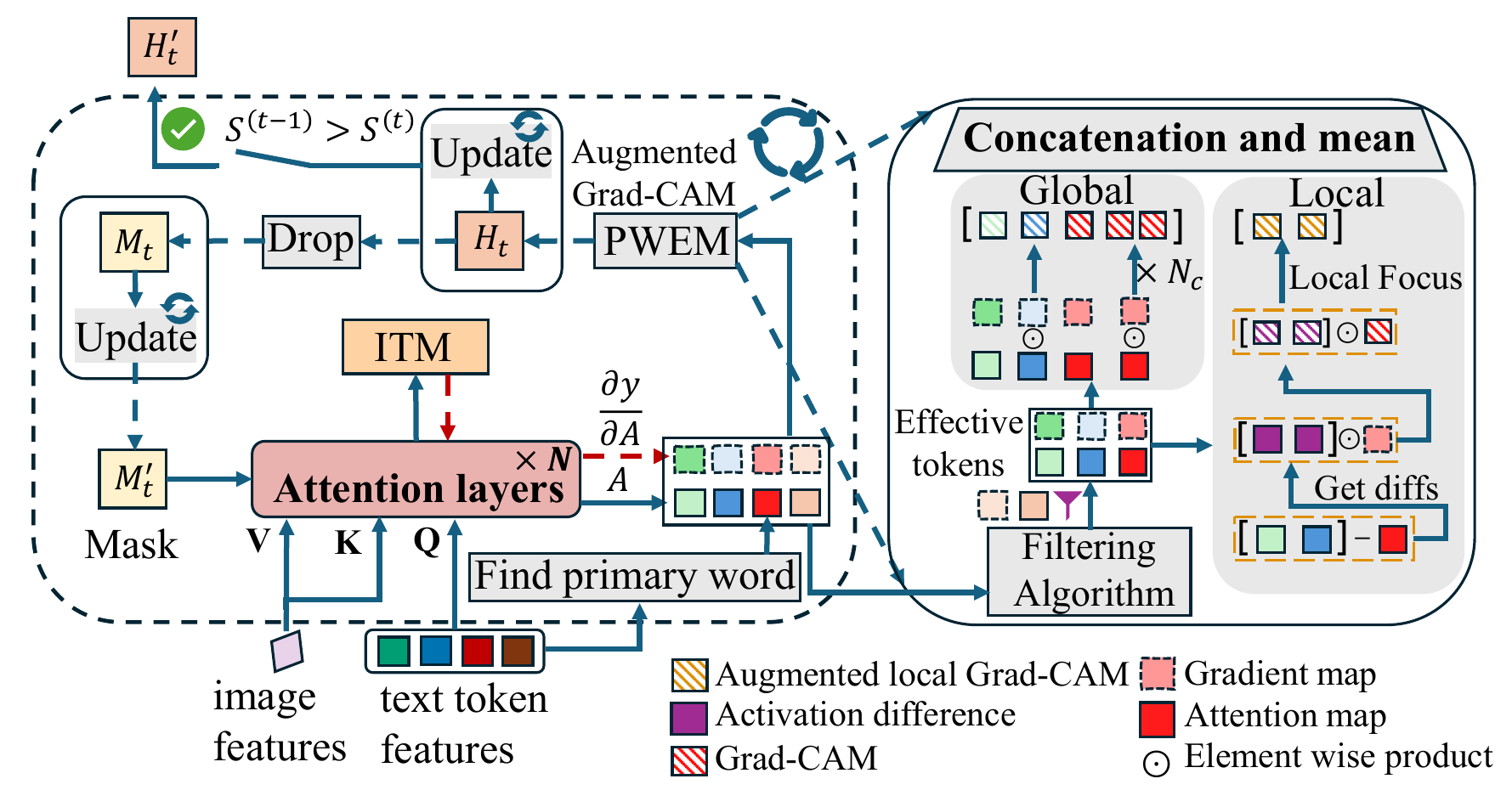}
	\caption{The proposed IGRS (left) and PWEM (right). The mask $M_{t}^{'}$ is the attention mask for cross-attention layers by dropping the most salient regions of Grad-CAM to zero. PWEM filters the meaningless tokens and augments the Grad-CAM representation from local and global aspects. }
	\label{fig3}
\end{figure*}
\subsection{Iterative Grad-CAM Refinement Strategy}
Masked Language Modeling (MLM) can be used for bi-directional image generative Transformers such as MasGIT \cite{maskgit}. The iterative generative paradigm offers self-correction chances for the model to optimize step-by-step in the latent space. Inspired by this, we propose a novel iterative strategy of Grad-CAM to gradually steer the model's attention to the region that the model is not attentive to initially, which brings benefits from two sides. On the one hand, for the circumstances in which Grad-CAM correctly localizes the instance initially, the gradually refined Grad-CAM can be better gathered around the targeted instance region. On the other hand, for the first incorrect localization, the model can attend to other semantic instances to recheck the Grad-CAM prediction, especially for the positional phrase inputs. The overall approach of IGRS is illustrated in the left part of Figure \ref{fig3}.

For simple notification, we use ${H}_{t}$ to represent the resultant $t$-th iteration Grad-CAM from the PWEM $\mathbf{H}_{a}$. Equation (\ref{eq5}) delineates the aggregation and refinement process of Grad-CAM representational updation, which entails the combination with Grad-CAM in the penultimate iteration step $(t-1)$, under the constraint of a zero initial condition $H_{0}=0$.
\begin{equation}\label{eq5}
    {H}_{t}^{'} = \lambda {H}_{t-1}^{'} + (1 - \lambda) \sigma({H}_{t}),
\end{equation}
where ${H}_{t-1}^{'}$ and ${H}_{t}^{'}$ are the resultant refined heatmaps from the $(t-1)$-th and $t$-th iterations,  $\sigma(.)$ is a sigmoid function to scale the value appropriately, and the hyperparameter $\lambda$ is a balancing factor. To instruct the model to focus on the region previously not paid attention to, in each iteration, a binary attention mask $M_{t}$ would be generated from the refined Grad-CAM heatmap ${H}_{t}$ by dropping the most attentive region to 0, as shown in Equation (\ref{eq6}).
\begin{equation}\label{eq6}
    M_{t} = \mathcal{P}({H}_{t}, \theta), \mathcal{P}({H}, \theta) = 
\begin{cases} 
0 & \text{if } \sigma({H}) \geq \theta \\
1 & \text{if } \sigma({H}) < \theta ,
\end{cases}
\end{equation}
where $\mathcal{P}({H}, \theta)$ represents the process of applying a sigmoid function to stretch the values and then thresholding the result at $\theta$ to create a binary mask. The binary mask $M_{t}$ is then combined with the previous mask $M_{t-1}^{'}$ by the logical \textit{and} operation, $M_{t}^{'} = M_{t-1}^{'} \land M_{t}$, where $M_{0}$ is a tensor of ones. The $\land$ ensures the model can expand to other regions regardless of the places previously focused. This attention binary mask will be fed into the cross-attention layer of a VLP to mask out the visual regions in embedding $v$, ensuring the text token queries no longer pay attention to the zero regions within the mask. 

For an interactive algorithm, the stopping condition is essential. To make the iterative process more flexible, we introduce a dynamic stopping criterion based on the proposed soft ITM score at timestep $t$, ${S}^{(t)}$, which is calculated as the product of the ITM, from the VLP model and the relevance score, ${S}^{(t)} = {ITM}^{(t)} \cdot R^{(t)}$, where $R^{(t)}$ is defined by:
\begin{equation}
    R^{(t)} = \frac{\sum\nolimits_{x \in X} {\sum\nolimits_{y \in Y} {(1 - {\widetilde{H}}_{t-1}^{'}})} }{{X \times Y}},
\end{equation} 
where ${\widetilde{H}}_{t-1}^{'}$ is the interpolated Grad-CAM heatmap of ${H}_{t-1}^{'}$ with the same size as the original image with width $X$ and height $Y$. This relevance score measures the average over-looked Grad-CAM intensity. A higher $R^{(t)}$ indicates that there are some regions less attentive to previously, guiding the model to focus on these overlooked areas in the next iteration. If the score for the current iteration ${S}^{(t)}$ is less than the score from the previous iteration $S^{(t-1)}$, the iterative process is terminated. The total iterative times should not exceed $\nu $.

\begin{table*}[t]
    \centering
    \scalebox{0.88}{
        \begin{tabular}{c|c|c|c|c|c|c|c|c|c|c|c|c}
            \toprule
            \multirow{2}{*}{Methods} & \multicolumn{4}{c|}{RefCOCO} & \multicolumn{4}{c|}{RefCOCO+} & \multicolumn{3}{c|}{RefCOCOg} & \multirow{2}{*}{Average} \\ \cline{2-12}
            & val & testA & testB & avg. & val & testA & testB & avg. & val & test & avg. & \\ \hline
            \textit{Zero-shot methods} & & & & & & & & & & & & \\ 
            GL-CLIP \cite{gcclip} & 26.7 & 25.0 & 26.5 &26.1 & 28.2 & 26.5 & 27.9 &27.5 & 33.0 & 33.1 &33.1 & 28.4 \\ 
            BSAP \cite{bsap} & 27.3 & 27.0 & 27.1 &27.1 & 28.7 & 27.8 & 28.3 &28.3 & 34.5 & 34.5 &34.5 & 29.4 \\ 
            Region token \cite{gcclip} & 23.4 & 22.1 & 24.6 &23.4 & 24.5 & 22.6 & 25.4 &24.2 & 27.6 & 27.3 &27.5 & 24.7 \\ 
            SAM-CLIP \cite{refdiff} & 26.3 & 25.8 & 26.4 &26.2 & 25.7 & 28 & 26.8 &26.8 & 38.8 & 38.9 &38.9 & 29.6 \\ 
            Ref-Diff \cite{refdiff} & 37.2 & 38.4 & \textbf{37.2} &37.6 & 37.3 & 40.5 & 33 &36.9 & 44 & 44.5 &44.3 & 39.0\\ 
            TAS \cite{tas} & 39.8 & 41.1 & 36.2 &39.0 & 43.6 & 49.1 & \textbf{36.5} &43.1 & \textbf{46.6} & \textbf{46.8} &\textbf{46.7} & 42.5 \\
            CaR \cite{car} & 33.6 & 35.4 & 30.5 &33.0 & 34.2 & 36.0 & 31.0 &33.7 & 36.7 & 36.6 &36.7 & 34.3 \\ \hline
            \textit{Weakly-supervised methods} & & & & & & & & & & & & \\ 
            TSEG \cite{tseg} & 25.4 & - & - & -& 22.0 & - & - &- & 22.1 & - &- & -\\ 
            Chunk \cite{chunck} & 31.1 & 32.3 & 30.1 &31.8 & 31.3 & 32.1 & 30.1 &31.2 & 32.9 & - &- & - \\ \midrule
            IteRPrimE (ours) & \textbf{40.2} & \textbf{46.5} & 33.9 &\textbf{40.2}
 & \textbf{44.2} & \textbf{51.6} & 35.3 &\textbf{43.7} & 46.0 & 45.8 &45.9 & \textbf{42.9} \\ \bottomrule
        \end{tabular}
    }
    \caption{Comparison of different methods on different datasets. ``avg." denotes the mean performance across various splits within individual datasets, while the terminal ``Average” column represents the composite mean derived from all dataset splits.}
    \label{tab: refcoco sota}
\end{table*}

\subsection{Selective Mask Proposal Network}
Through the aforementioned steps, we can employ the Grad-CAM indication clue to instruct the Segmentors to predict the referred instance mask. For a given image, the mask proposal network would predict the $N_b$ masks but they can not autonomously choose which object mask users refer to by the language. Therefore, the selection module within the network is designed to select the mask indicated by the Grad-CAM, which divides the selection procedures into two phases: the filtering phase and the scoring phase.

Assuming that the Grad-CAM has successfully localized the instance, the center point of Grad-CAM should be within the inner part of the object. Based on this hypothesis, the selection mechanism is initiated by a preliminary evaluation that involves two main criteria. First, we identify the set of coordinates, $\mathcal{C}_{{max}}$, where the heatmap reaches its peak values. Then, the $m$-th candidate mask ${B}^{m}$ is examined to determine if it includes at least one activated pixel at any of these coordinates. Second, to ensure the quality of the masks, we apply a connected component labeling technique to constrain the number of connected components in each mask, ensuring that the number of these components does not exceed a predefined threshold of $\kappa$. The combined criteria for the preliminary evaluation are defined as follows:
\begin{equation}
\label{eq:selection_criteria}
\begin{aligned}
\mathcal{A} &= \left\{  m \in {N_b} \mid {B}_{(x,y)}^{m} \neq 0, \exists (x,y) \in \mathcal{C}_{{max}}, \right\}, \\
\mathcal{F} &= \left\{ m \in {N_b}\mid g_{cc}({B}^{m}) \leq \kappa \right\}, \\
\mathcal{D} &= \mathcal{A} \cap \mathcal{F}.
\end{aligned}
\end{equation}
In the above equations, $g_{cc}$ denotes a function that quantifies the number of connected components within the $m$-th candidate from total $N_b$ masks. The intersection of sets $\mathcal{A}$ and $\mathcal{F}$, denoted as $\mathcal{D}$, yields the subset of candidate masks that fulfill both the activation and mask quality requirements. This evaluation process filers the irrelevant and empty masks to reduce the computational cost and enhance efficiency.

Subsequent to the preliminary filtering phase, we proceed to evaluate each remaining candidate mask through a weighted scoring mechanism that leverages the Grad-CAM heatmap. This involves computing an element-wise product-based score for each mask concerning the heatmap. We define the score for the $j$-th candidate mask as $Z(j)$ from the set $\mathcal{D}$. The scoring process is formulated below:
\begin{equation}
\label{eq:scoring_process}
\begin{aligned}
{Z}(j) &= \sum\limits_{x \in X} {\sum\limits_{y \in Y}  {\left({B}_{(x,y)}^{j} + {B}_{(x,y)}^{j} \odot \widetilde{{H}}_{(x,y)}^{'}\right)}}  \\
\hat Z(j){\rm{ }} &= {\rm{ }}\frac{{Z(j)}}{{\sum\nolimits_{x \in X} {\sum\nolimits_{y \in Y} {B_{_{(x,y)}}^j} } }}, \quad j \in \mathcal{D}.
\end{aligned}
\end{equation}
where $\widetilde{{H}}_{(x,y)}^{'}$ is the final output Grad-CAM of original image size. The final step in our selection process involves identifying the candidate mask with the maximum normalized score, $\hat Z(j)$, as the chosen segmentation output: 
\begin{equation}
    B_{select} = \arg\max_{j \in \mathcal{D}} \hat Z(j).
\end{equation}
This approach ensures that the selected mask aligns with the regions of interest highlighted by the Grad-CAM heatmap, thereby ensuring the precision and efficacy of RIS.

\begin{table}[t]
\centering
\scalebox{1}{\begin{tabular}{c|c|c|c}
\toprule
\textbf{Method} & \textbf{Training dataset} & \textbf{All} & \textbf{Unseen} \\ \hline
\multirow{3}{*}{CRIS} & RefCOCO & 15.5 & 13.8 \\ 
& RefCOCO+ & 16.3 & 14.6 \\ 
& RefCOCOg & 16.2 & 13.9 \\ \hline
\multirow{3}{*}{LAVT} & RefCOCO & 16.7 & 14.4 \\ 
& RefCOCO+ & 16.6 & 13.5 \\ 
& RefCOCOg & 16.1 & 13.5 \\ \hline
GL-CLIP & N/A & 23.6 & 23.0 \\ 
TAS & N/A & 25.6 & - \\ 
Ref-Diff & N/A & 29.4 & - \\ \hline
IteRPrimE (ours) & N/A & \textbf{38.1} & \textbf{37.9} \\ \bottomrule
\end{tabular}}
\caption{Comparison of oIoU on PhraseCut for different supervised and zero-shot methods.}
\label{tab:phrasecut}
\end{table}

\begin{figure*}[t]
	\centering	\includegraphics[width=0.8\textwidth]{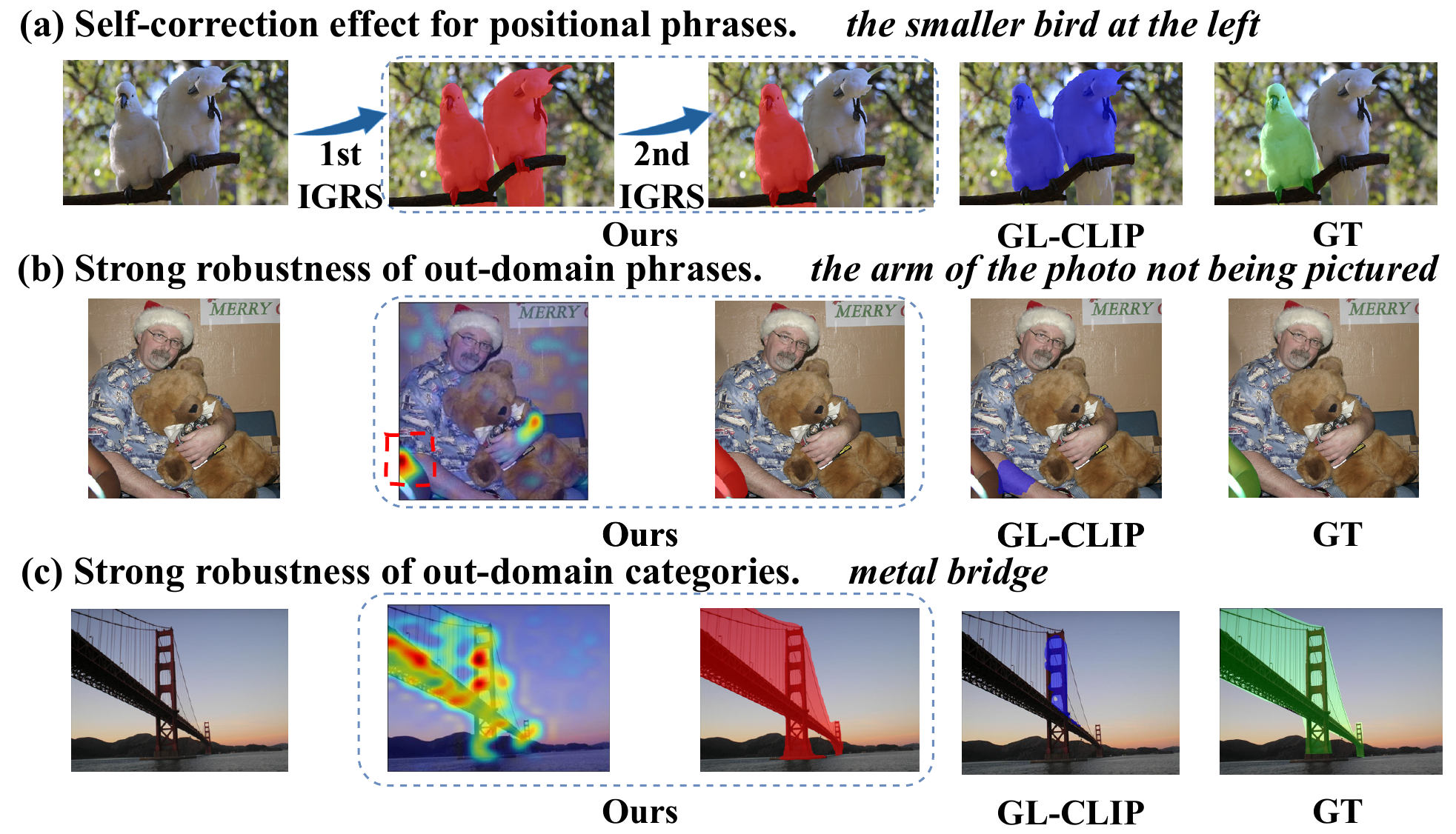}
	\caption{The qualitative comparisons with GL-CLIP. (a) The self-correction effect is brought by our IGRS, especially for positional phrases. (b) For the unseen phrases like ``not", our model shows better robustness. (c) shows the gathering effect of IteRPrimE with high confidence to select the whole mask instead of a part like GL-CLIP.    }
	\label{fig:fig4}
\end{figure*}

\section{Experiments}

% In this section, we describe the implementation and experimental results in detail. 
\subsection{Experimental Settings}
{\bf Datasets and metrics.} We employ the RefCOCO \cite{refcoco}, RefCOCO+ \cite{refcoco}, RefCOCOg \cite{refcocog1, refcocog2}, and PhraseCut datasets \cite{phrasecut} for evaluating the proposed zero-shot methods. RefCOCO with shorter expressions (average 1.6 nouns, 3.6 words) contains massive positional phrases (50\%), especially those with direct direction clues like ``left" or ``right". In contrast, RefCOCO+ focuses on the attribute phrases with the same average expression length. RefCOCOg is a more challenging benchmark that has longer phrases (average 2.8 nouns, 8.4 words) and complex expressions. To verify the effectiveness of the mode in out-of-domains, we adapt our model to the PhraseCut dataset which contains the additional 1271 categories in the test split based on 80 in COCO. Following \cite{car,gcclip,zerorec}, we utilize the mean Intersection over Union (mIoU) for RefCOCO series, a common metric for RIS. Following \cite{gcclip, phrasecut}, we report the overall Intersection over Union (oIoU) for the PhraseCut dataset.

{\bf Implementation details.} We use the commonly used mask proposal network, Mask2Former \cite{mask2former, ovseg}, to obtain 200 instance-level mask proposals. Following \cite{groundvlp, chunck}, we utilize the base model ALBEF to study the Grad-CAM for localization and it is generated in the 8th cross-attention layer. In processing the input text, a prefatory phrase “there is a” is appended. The hyperparameter balancing factor $\lambda$, upper connecting limit $\kappa $, iterative number $\nu$, and binarization threshold $\theta $ are 0.8, 12, 3, and 0.5, respectively. All experiments are conducted on a 24 GB RTX 3090 GPU.

\begin{table}[t]
\centering
\scalebox{0.88}{\begin{tabular}{c|c|c}
\toprule
\textbf{Method} & \textbf{RefCOCO testA} & \textbf{RefCOCOg test} \\ \hline
Overall Mean & 43.4 & 41.3 \\
GVLP \cite{groundvlp} & 45.0 & 41.9 \\ \hline
Global Augment & \textbf{46.5} & 45.7 \\
Local Augment & 45.3 & 42.3 \\
PWEM & \textbf{46.5} & \textbf{45.8} \\
\bottomrule
\end{tabular}}
\caption{Comparison of methods with different Grad-CAM generation methods on RefCOCO testA and RefCOCOg test datasets.}
\label{tab:PWEM}
\end{table}

\begin{table}[t]
\centering
\scalebox{0.8}{\begin{tabular}{c|c|c|c}
\toprule
\textbf{Method} & \textbf{RefCOCO testA} & \textbf{RefCOCOg test} & \textbf{RefCOCOg val} \\ \hline
Mask image & 46.3 & 45.3 & 45.2 \\
Mask feature & \textbf{46.5} & \textbf{45.8} & \textbf{46.0} \\
\bottomrule
\end{tabular}}
\caption{Performance comparison of masking out the salient regions in the image level and feature level (attention mask).}
\label{tab:shallow_deep_comparison}
\end{table}

\subsection{Results}

% In this section, we present the experimental results of our prompt counting method and compare its performance against other class-agnostic counting methods. 
{\bf Main results.} As shown in Table~\ref{tab: refcoco sota}, IteRPrimE almost achieves the best performance on all three datasets, especially in the testA splits of RefCOCO and RefCOCO+. It outperforms the SOTA TAS method with a 0.4\% average improvement. For all the splits of RefCOCO and RefCOCO+ rich in short positional phrases, our model obtains an average of 40.2\% and 43.7\% compared to the 39.0\% and 43.1\% of TAS, respectively. Therefore, our method is more robust to the positional information compared to the CLIP-based paradigms. However, the model may have the relatively weaker capability of complex expressions shown in RefCOCOg, which can be attributed to the data limitation and gap in the pertaining stage. Additionally, by using the additional captioner of BLIP2 \cite{blip2} and SAM \cite{sam}, TAS maintains the best performance across some splits, especially for complex phrases, but it has the drawback of low throughput and heavy volumes. 

{\bf Zero-shot evaluation on unseen domain.} Notably, as shown in Table \ref{tab:phrasecut}, our model has high capabilities of cross-domain zero-shot transfer compared to other zero-shot SOTA and the existing supervised methods CRIS \cite{cris} and LAVT \cite{lavt}. IteRPrimE significantly outperforms both kinds of methods in the out-domain scenarios. Upon assessment within a subset of categories not present in the RefCOCO datasets (denoted as the ``Unseen" column), our model shows the best robustness compared to the supervised methods with huge performance degradation. Notably, the underperformance of the TAS model on this dataset may be attributed to the predominance of complex outdoor scenes within the dataset. In such intricate environments, the reliance on an additional captioning model for annotation by TAS could potentially introduce greater noise, thereby compromising the model’s performance. However, facing complex environmental contexts, our model's efficacy in localizing pertinent regions is attributed to its retention of spatial perception. Concurrently, the integration of IGRS and PWEM has further bolstered IteRPrimE’s proficiency in addressing the complicated interrelationships among objects within the scene, thereby leading to this commendable performance.

{\bf Qualitative comparisons.} Figure \ref{fig:fig4} shows the comparisons with GL-CLIP \cite{gcclip}. First, we demonstrate that our IGRS module possesses a self-corrective mechanism, the same answer as GL-CLIP initially before refining its predictions by revisiting initially overlooked regions. In Figure \ref{fig:fig4} (b), the scarcity of such negative phrases in the training set is offset by our model’s robustness. Finally, we address the limited highlighted region of initial Grad-CAM representation by the IGRS, demonstrated in Figure \ref{fig:fig4} (c). The more gathering of the Grad-CAM, the more likelihood that the correct instance mask will be selected instead of the part.

\subsection{Ablation Study}
{\bf Effect of PWEM.} According to Equation (\ref{eq2}), the mean operation is essential for the generation of Grad-CAM and deeply influences the Grad-CAM representational accuracy. Therefore, Table \ref{tab:PWEM} presents the results of the ablation study, examining the impact of various aggregation configurations for Grad-CAM generation. The ``Overall Mean" is the direct mean of all the tokens' Grad-CAM, but the GVLP uses the selected effective tokens for averaging \cite{groundvlp}. The remaining is introduced before as shown in Equation (\ref{eq3}) and Equation (\ref{eq4}).  Compared to the previous methods, the proposed PWEM can significantly improve the performances because it can save the examples that fail due to the weak complex semantic understanding between the main word and the other contexts. Additionally, global augmentation shows stronger potential than local because it could dominate the effect during aggregation.

{\bf Effect of mask position in IGRS.} Table \ref{tab:shallow_deep_comparison} evaluates the position that the binary mask $M$ applied. ``Mask image'' means adding the mask into the original image so that the indicated regions are masked out, similar to GL-CLIP. However, this can degrade the performance due to the absence of relative relationships of regions. Our method for attention masking in the cross-attention layer is more robust, with improvement on all three splits.

% \begin{table}[t]
% \centering
% \scalebox{0.6}{\begin{tabular}{c|c|c|c|c|c|c|c|c|c}
% \toprule
% \multirow{2}{*}{\textbf{Method}} & \multicolumn{3}{c|}{\textbf{RefCOCOg test}} & \multicolumn{3}{c|}{\textbf{RefCOCO testA}} & \multicolumn{3}{c}{\textbf{RefCOCOg val}} \\ \cline{2-10}
% & Position & Others & Overall & Position & Others & Overall & Position & Others & Overall \\ \hline
% GVLP w/o Iter & 33.0 & 43.6 & 41.3 & 34.7 & 53.2 & 44.7 & 34.2 & 44.3 & 42.0 \\
% GVLP w/ Iter & 33.7 & 44.3 & 41.9 & 35.1 & 53.6 & 45.0 & 34.5 & 44.7 & 42.4 \\
% PWEM w/o Iter & 36.4 & 47.5 & 45.1 & 36.1 & 54.8 & 46.1 & 37.6 & 47.4 & 45.1 \\
% PWEM w/ Iter & \textbf{37.4} & \textbf{48.2} & \textbf{45.8} & \textbf{36.5} & \textbf{55.0} & \textbf{46.5} & \textbf{38.3} & \textbf{48.2} & \textbf{46.0} \\

% \bottomrule
% \end{tabular}}
% \caption{Comparison of methods across different datasets and positions.}
% \label{tab:comparison}
% \end{table}

\begin{table}[t]
\centering
\scalebox{0.7}{\begin{tabular}{c|c|c|c|c|c|c}
\toprule
\multirow{2}{*}{\textbf{Method}} & \multicolumn{3}{c|}{\textbf{RefCOCOg test}} & \multicolumn{3}{c}{\textbf{RefCOCO testA}} \\ \cline{2-7}
& Position & Others & Overall & Position & Others & Overall \\ \hline
GVLP w/o IGRS & 33.0 & 43.6 & 41.3 & 34.7 & 53.2 & 44.7 \\
GVLP w/ IGRS & 33.7 & 44.3 & 41.9 & 35.1 & 53.6 & 45.0 \\
PWEM w/o IGRS & 36.4 & 47.5 & 45.1 & 36.1 & 54.8 & 46.1 \\
PWEM w/ IGRS & \textbf{37.4} & \textbf{48.2} & \textbf{45.8} & \textbf{36.5} & \textbf{55.0} & \textbf{46.5} \\

\bottomrule
\end{tabular}}
\caption{Ablation studies of the proposed PWEM and IGRS.}
\label{tab:comparison}
\end{table}

{\bf Effect of our proposed PWEM and IGRS.} Table \ref{tab:comparison} evaluates the performance improvements achieved by integrating different modules within our methodology. The ``Position" category encompasses those test samples that explicitly feature positional expressions. Conversely, the ``Others" category serves as the complement. These results demonstrate that our modules not only improve general performance but also enhance the model's ability to manage complex semantic and spatial relations, particularly in positional contexts.

\begin{figure}
	\centering	\includegraphics[width=0.26\textwidth]{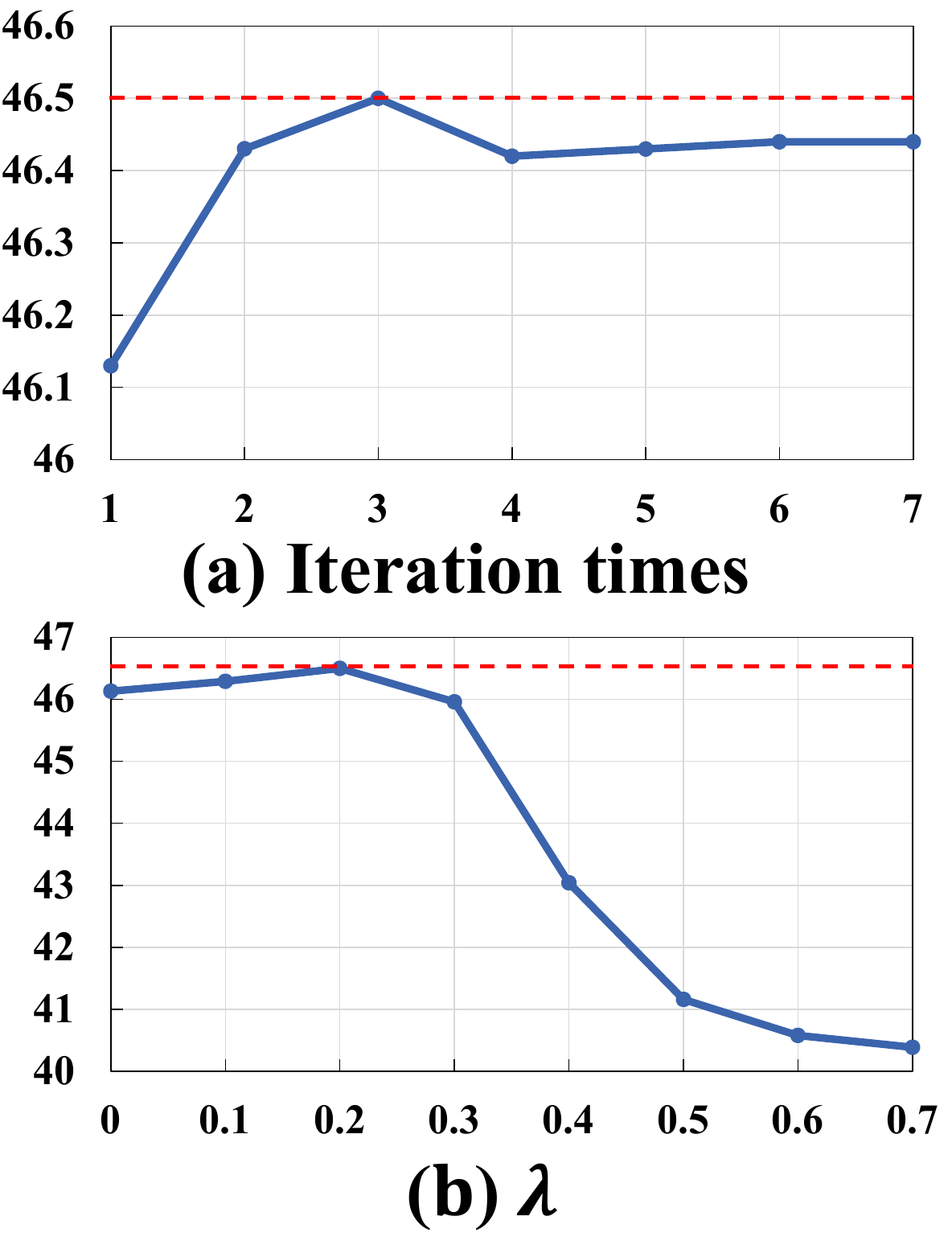}
	\caption{The line charts of two hyperparameters. }
	\label{fig:hyperparameter}
\end{figure}

{\bf Different assembly of iteration times and $\lambda$.} Figure \ref{fig:hyperparameter} presents the ablation study of two hyperparameters in IGRS, analyzing the effect of varying iteration times and $\lambda$ on the RefCOCO testA dataset. Figure \ref{fig:hyperparameter} (a) shows that as the number of iterations increases from 1 to 3, the metric improves, peaking at 46.5\%. However, beyond three iterations, the performance change becomes minimal. Therefore, selecting 3 iterations is optimal for balancing performance and time efficiency. Figure \ref{fig:hyperparameter} (b) presents another line chart analyzing the impact of $\lambda$ in the Grad-CAM updation. The metric increases as $\lambda$ is gradually raised from 0 to 0.2 while exceeding this point, performance declines with higher alpha values. Overall, the optimal value of $\lambda$ is 0.2.

\section{Conclusion}
This paper presents IteRPrimE, a novel framework for Zero-shot Referring Image Segmentation (RIS), addressing the limitations of previous methods in handling positional sensitivity and complex semantic relationships. By incorporating an Iterative Grad-CAM Refinement Strategy (IGRS) and a Primary Word Emphasis Module (PWEM), IteRPrimE enhances the model’s ability to accurately focus on target regions and manage semantic nuances. Extensive experiments on RefCOCO/+/g and PhraseCut benchmarks demonstrate that IteRPrimE significantly outperforms previous state-of-the-art zero-shot methods, particularly in out-of-domain contexts. These findings highlight the framework's potential to advance zero-shot RIS by improving model sensitivity to positional and semantic details. Future research endeavors may seek to extend the Grad-CAM-guided RIS paradigm to encompass all segmentation tasks across varying levels of granularity with linguistic directives. 

\section{Acknowledgments} 

This work was supported by Shenzhen Science and Technology Program under Grant CJGJZD20220517142402006.

\bibliography{aaai25}

\end{document}